# DeepGuard: A Framework for Safeguarding Autonomous Driving Systems from Inconsistent Behaviours

[1]Manzoor Hussain[0000-0002-9131-0930], Nazakat Ali[0000-0002-3875-812X], and *Jang Eui-Hong[0000-0001-9786-7732]

*Abstract*— Safety assurance is vital to the deep neural network (DNN)–based autonomous driving systems (ADSs). However, DNN-based ADS is a highly complex system that puts forward a strong demand for robustness, more specifically, the ability to predict unexpected driving conditions to prevent potential inconsistent behaviour. It is not possible to generalize the DNN for all driving conditions. Therefore, the driving conditions that were not considered during the training of the ADS may lead to unpredictable consequences for the safety of autonomous vehicles. This study proposes an autoencoder and time series analysis–based anomaly detection system to prevent the safety-critical inconsistent behaviour of autonomous vehicles at runtime. Our approach called *DeepGuard* consists of two components. The first component- the inconsistent behaviour predictor, is based on an autoencoder and time series analysis to reconstruct the driving scenarios. Based on reconstruction error and threshold, it determines the normal and unexpected driving scenarios and predicts potential inconsistent behaviour. The second component provides on-the-fly safety guards, that is, it automatically activates healing strategies to prevent inconsistencies in the behaviour. We evaluated the performance of *DeepGuard* in predicting the injected anomalous driving scenarios using already available open-sourced DNN-based ADSs in the Udacity simulator. Our simulation results show that the best variant of *DeepGuard* can predict up to 93 % on the *CHAUFFEUR* ADS, 83 % on *DAVE-2* ADS, and 80 % of inconsistent behaviour on the *EPOCH* ADS model, outperforming *SELFORACLE* and *DeepRoad*. Overall, *DeepGuard* can prevent 89% of all predicted inconsistent behaviours of ADS by executing predefined safety guards.

## I. INTRODUCTION

Autonomous vehicles are one of the most promising applications of artificial intelligence. This would be a technological revolution in the transportation industry in the near future. Autonomous driving systems (ADSs) use sensors such as cameras, radar, Lidar, and GPS to automatically produce driving parameters such as vehicle velocity, throttle, brakes, steering angles, and directions. Advancements in deep learning have made progress in autonomous systems, such as autonomous vehicles and unmanned aerial vehicles. However, autonomous vehicles have attracted significant attention owing to their impact on economic effects and safety. Many enterprises and scientific research centers have developed advanced autonomous driving system models. Google, Tesla, Nvidia, and Apple developed their own versions of ADSs (Greenblatt 2016). Typical ADSs take data from the surroundings using sensors and the deep neural network (DNN) is used to process the data and produce the driving parameters.

Recent advancements in DNNs have increased the robustness of ADSs, and ADSs can adapt their driving behaviour in response to dynamic environments. Many end-to-end supervised learning frameworks are available to train the DNN-based ADSs for predicting driving behaviours, such as steering angles, by taking the driving images as input and using the image and driving

Manzoor Hussain
hussain@ selab.cbnu.ac.kr
Nazakat Ali
nazakatali@ selab.cbnu.ac.kr
Jang Eui Hong
Correspondnce Author: e-mail: jehong@ chungbuk.ac.kr
[1] School of Electrical and Computer Engineering, Chungbuk National University, Korea, Chung-daero, Seowon-gu, Cheongju-si, Chungbuk, South Korea

behaviour pair as training data. For example, NVIDIA's DAVE-2 can predict the steering angles based on only single images coming from the front camera (Bojarski et al. 2016).

Testing the ADS is a very challenging task. Autonomous car manufacturing companies perform only limited in-field testing within limited testing facilities and gather driving data during testing. Such expensive in-field testing cannot be afforded by manufacturing companies. Hence, this limited testing cannot cover all the driving scenarios. Therefore, it is unrealistic to generalize that the DNN model can perform well in all driving contexts. For these reasons, the manufacturer collects the sensor data of the vehicle during limited infield testing to recreate the comprehensive testing scenarios in the simulation environments. Simulators allow testing of the ADS under large-scale nominal conditions as well as dynamic environments, such as extreme weather conditions and dangerous driving scenarios (Cerf 2018).

Owing to a plethora of driving conditions, it is difficult to generalize that the DNN-based ADS will work under all driving conditions. Because unexpected driving contexts are by definition unknown during the training time; known conditions at training time would help to train the DNN better (Campos et al. 2016). Therefore, a system that may predict the potential inconsistencies in ADS behaviour is required when it enters an unexpected driving zone that was not considered during the training. Based on the severity of potential inconsistencies, the inconsistent behaviour prevention system may activate safety guards to bring the autonomous vehicle to a safe zone.

In this paper, we introduce an autoencoder and time series analysis–based approach to predict the potential inconsistencies in ADS due to unexpected driving contexts. Our proposed approach is called *DeepGuard*, which is based on a reconstruction-based technique to monitor the driving behaviour of ADS. Based on the different levels of reconstructed error, different levels of required safety guards are applied to ensure the safety of the ADS. *DeepGuard* observes the increasing trends in reconstruction error over time, which can tackle the potential inconsistencies in the behaviour of ADS by detecting unexpected driving conditions and by activating safety guards such as a *decrease in speed*, *applying the emergency brake*, and *safe disengagement of autonomous mode* before the violation of the safety properties. We used the Udacity simulator for autonomous cars (Udacity 2017) to evaluate the performance of our proposed approach using open-sourced ADS models. To create unexpected driving conditions (*such as raining driving conditions, dense fog, heavy snowfall, and snow on the road*) in our simulation, we used a modified version of the original Udacity simulator developed by *SELFORACLE* (Stocco et al. 2020). The best-performing variant of *DeepGuard* can predict 93% of the inconsistent behaviours. In our comparative analysis, we found that the *DeepGuard* approach outperforms *SELFORACLE* and *DeepRoad* (Zhang et al. 2018) in terms of model performance as well as the safety-guard approach to ensure safety of the ADS. In this paper we make the following contributions:

1. **Enhanced Inconsistent Behaviour Predictor:** We proposed an unsupervised technique to predict the potential inconsistent behaviour of the ADS based on autoencoders and time series analysis.
2. **Inconsistent Behaviour Prevention System:**
    a. We designed generalized safety guards for DNN-based ADSs for unexpected driving contexts.
    b. We introduced an inconsistent behaviour prevention system by activating safety guards at runtime to ensure the safety of ADS in unexpected driving contexts.
3. **Comparative Analysis:** We compared our approach with *SELFORACLE* and *DeepRoad* to validate its effectiveness and performance. Our simulation results showed that *DeepGuard* outperforms *SELFORACLE* and *DeepRoad* in terms of prediction and prevention of inconsistent ADS behaviour.

The remainder of this paper is organized as follows. Section II provides related works and the limitations of existing approaches. Section III explains the prevention of inconsistent behaviours in the autonomous driving system by executing safety guards at runtime in detail. The experimental evaluation is presented in Section IV. Finally, the conclusion is laid out in Section V.

## II. Related Work

Autonomous vehicles are safety-critical systems, especially on highways in adverse weather conditions such as dense fog, heavy rain, and snow falling, where the speed of the vehicles is usually very high. In such driving contexts, ensuring the safety of autonomous vehicles is a challenging task. The common approach to achieve safe driving in such a dangerous situation is path following and yaw stability. Path following can be achieved by utilizing predictive model control, and yaw stability can be achieved with the help of supplementary yaw moment generated from the brake and active steering combined with the brake control (Tjonnas and Johansen 2010). Such an approach was presented by (Zhang et al. 2020) to evaluate the applicability of model predictive control (MPC) and torque vectoring to path following and yaw stability control of over-actuated autonomous electric vehicles during safety-critical driving conditions.

Autoencoders are widely used in anomaly detection systems in different domains (Borghesi et al. 2019; Chen et al. 2018; Cozzolino and Verdoliva 2016; Zavrtanik et al. 2021). The authors (An and Cho 2015) proposed a variational autoencoder-based anomaly detection system based on the reconstruction probability. The authors in (Bahavan et al. 2020) proposed a self-supervised deep learning algorithm to detect anomalies in autonomous systems based on a long short-term memory (LSTM) autoencoder. The received frame is set to be anomalous or normal based on the prediction errors, which is the degree of difference estimated using the reconstruction error and threshold. In another study, the authors (Azzalini et al. 2021) used a variational autoencoder to detect anomalies in autonomous robot systems. In their study, the authors introduced a minimally supervised technique based on variational autoencoders.

In a platooning system, vehicles are supposed to drive by maintaining a very short distance among the vehicles (Bermad et al. 2019). Therefore, detecting anomalous and normal information is very important for connected vehicles. The authors (Nguyen et al. 2020) proposed an approach to enhance misbehaviour detection in 5G vehicle-to-vehicle communication. In their approach, the system verifies the behaviour of the preceding vehicle and then verifies the reliability of the received data during the cooperative adaptive cruise control (CACC) mode. In another study, the authors (Gyawali et al. 2020) proposed a misbehaviour detection system in vehicular communication networks using a machine-learning approach.

According to the global status report of the World Health Organization (WHO) on road safety in 2018, annual traffic accidents have reached 1.35 million (World Health Organization 2018). One of the main causes of traffic accidents is rear-ended collisions (Kusano and Gabler 2012). Generally, the forward collision occurs because of insufficient time to react to potential dangers and apply brake lately. According to a study by the authors (Wang et al. 2020), 90% of accidents can be avoided if the driver is warned up to 1.5 s in advance. The authors proposed a convolutional neural network (CNN)–based real-time collision prediction mechanism based on real trajectory datasets.

DNN has become fundamental to image-based ADSs. However, these systems may exhibit anomalous behaviour that may cause accidents, resulting in fatalities. An interesting approach called *SELFORACLE* (Stocco et al. 2020) was proposed to predict the potential misbehaviour in ADSs. In their approach, they used the concept of a self-assessment oracle approach using autoencoders and time series analysis, which monitors the confidence of DNN-based ADS at runtime. The authors used the concept of estimating the confidence of a DNN in response to an unexpected execution context to predict the potential misbehaviour and execute healing strategies at runtime. In another study, an approach called *DeepRoad* was proposed by (Zhang et al. 2018) to detect autonomous driving system inconsistencies. *DeepRoad* uses a generative adversarial network (GAN)–based metamorphic approach to generate driving scenes with various weather conditions to detect inconsistencies in ADSs.

The major issue with the machine learning-based autonomous system is to find the ability to distinguish between normal and anomalous behaviours. Autoencoders are currently used to detect anomalous and normal behaviour in many domains, such as surveillance videos (Chong and Tay 2017), robotics (Chen et al. 2020), and web applications (Xu et al. 2018). These approaches have been proposed for classification problems to detect misbehaviour. Several studies have been conducted to find unexpected execution scenarios based on anomaly detection systems, as mentioned in (Bolte et al. 2019; Patel et al. 2018). In another study, the authors (Henriksson et al. 2019) used the negative log of the reconstructed image by a variational autoencoder to detect anomalies in the driving images.

From the related works, we summarize the limitations of the existing approach as follows: 1) Apart from the *SELFORACLE* approach, none of these approaches considered the healing mechanism to ensure safety once the inconsistent behaviour is predicted. However, even in *SELFORACLE*, the healing system still does not provide sufficient healing for ADS because their self-healing system is activated after inconsistent behaviour has occurred. 2) No corrective actions or safety guard executions at runtime were considered in these approaches. The execution of safety guards is crucial for vehicle safety in the case of unexpected driving contexts. 3) Instead of relying on the healing of ADS after inconsistencies occurred, the execution of safety guards would be more effective in ensuring safety once the potential inconsistencies are predicted. In a real scenario, once misbehaviour occurs, the damage to the vehicle or even the safety of passengers is already compromised. Therefore, compared to the self-healing system approach, the execution of the safety guards approach would be better once unexpected driving conditions are detected to avoid inconsistent behaviour.

### III. Proposed Approach

ADSs have the potential to reduce accidents and injuries, as 94% of serious crashes are caused by human errors (NHTSA 2020). However, DNN-based driving systems sometimes exhibit inconsistent behaviours owing to anomalous driving contexts and cause fatal crashes. ADSs are error-prone to unexpected driving scenarios as mentioned in (Jana et al. 2018; Pei et al. 2017). The authors in (Pei et al. 2017) unveiled the inconsistent behaviours of DNN systems, while (Jana et al. 2018) disclosed the various anomalous driving behaviours of well-known open-source autonomous driving models.

In the context of our proposed approach, the meaning of inconsistent behaviour is a violation of the safety requirements. There are many potential inconsistent behaviours associated with DNN-based autonomous cars. For example, inconsistencies in the steering components may cause fatal crashes. However, in this study, we focus on the prevention of two safety-critical inconsistent behaviours of ADSs: 1) *vehicle collision* and 2) *lane departure* because these safety-critical inconsistent behaviours are critical for the safety of ADSs as well as humans. We aim to predict the potential inconsistent behaviours based on the increasing trends in the reconstruction error and preventing these inconsistent behaviours by executing safety guards before it occurs. To address the above-mentioned safety-critical issues, we present an inconsistent behaviour prevention system to ensure the safety of ADSs. The proposed approach consists of two components.

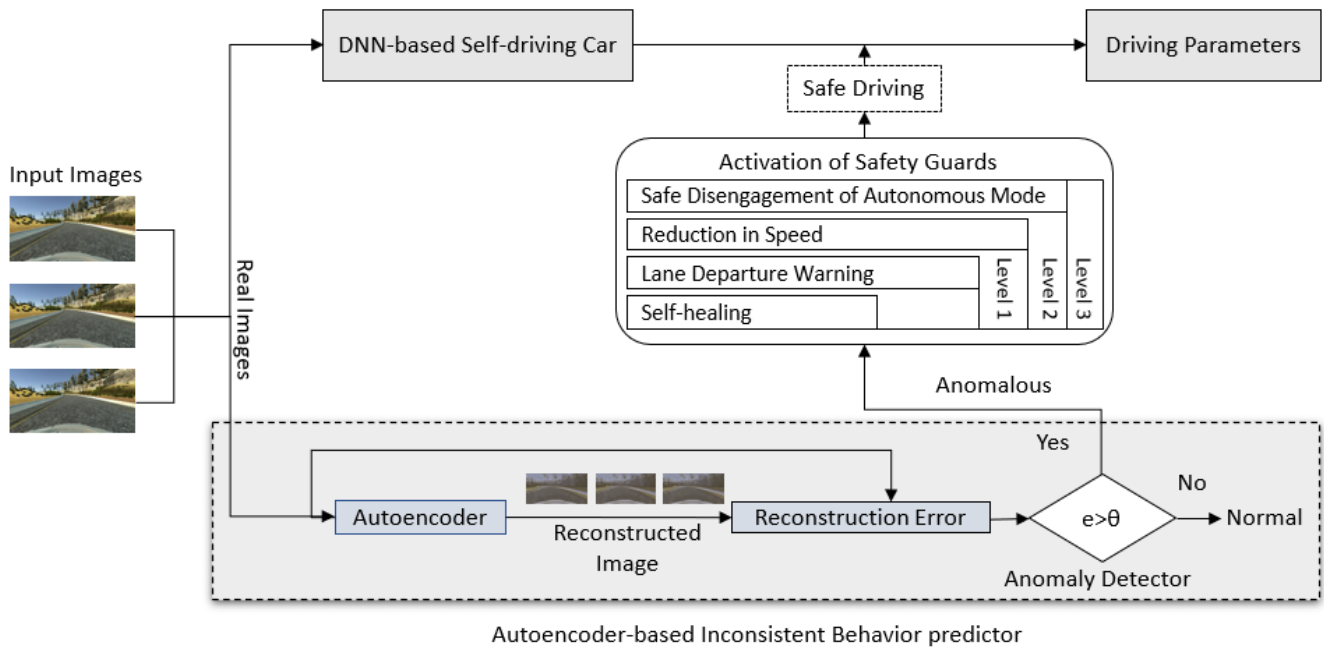

**Fig. 1** An approach to predict the inconsistent behaviour of autonomous driving systems and ensuring the safety of the ADS by executing safety gaurds at runtime.

**Part1: Inconsistent Behaviour Predictors:**

The inconsistent behaviour predictor of ADS is based on autoencoders that use the black-box technique. In the black-box technique, DNN uncertainty is predicted (i.e., in unexpected situations) by calculating the difference between the input images and the image data used for training the DNN. Let us take an example of an ADS, which was trained using the images collected during a sunny day on a highway. However, for the evaluation, the ADS was placed on a road where the driving scenario was heavy rain at night. In this situation the DNN may still produce some driving parameters; however, there is a chance of accidents as the driving scenario was not considered during training. Therefore, in such an unexpected driving context, we would like to warn the DNN to activate safety guards to prevent potential crashes. We chose the black-box technique because it is independent of the DNN-based autonomous driving model. We do not need to modify the original ADS model to integrate it.

**Part 2: Prevention of Inconsistent Behaviour:**

ADSs have a significant ability to mitigate the problem of daily accidents due to negligence and mistakes of human drivers (Anderson et al. 2016), and the performance of ADSs is far better than that of human drivers in terms of perception (e.g., blind spots), execution (e.g., faster and more precise control of steering and acceleration), and decision-making in some cases (e.g., more accurate driving maneuvers in complex driving scenarios). However, autonomous vehicles cannot mitigate all crashes. For example, in extreme weather conditions, unexpected driving conditions still create safety challenges for ADSs as well as human drivers. Autonomous vehicles may perform worse than human drivers in some complex driving scenarios, as mentioned in (Bonnefon et al. 2015). Autonomous vehicles may pose serious risks, such as crashes caused by cyber-attacks.

Modern ADSs have inbuilt technologies, such as avoiding drifting into adjacent lanes or unsafe lane changes and automatic brake systems. Other features such as lane-keeping systems, adaptive cruise control, traffic jam assistants, highway autopilots, and self-parking are already used in modern ADS. Levels 3 and 4 of ADS drive itself, but the driver must take control of the ADS in some circumstances. Despite these safety features, ADS still poses some threat to safety in complex and unexpected driving conditions. The explicit safety-guard approach provides safety in such situations to prevent hazardous scenarios.

The inconsistent behaviour predictor predicts inconsistencies in ADS behaviour, and the prevention system deals with the safety of the ADS once the inconsistent behaviour is predicted. Figure 1 illustrates the working of *DeepGuard*. Our approach analyzes the input images from three cameras (*left, right, and center cameras*) of the car. We first train the inconsistent behaviour predictor model using the normal driving conditions (*i.e., sunny condition*), and then the model deploys along with the ADS. When the ADS enters unexpected driving conditions (*i.e., rain, fog, and snowy driving scenarios*), the reconstruction error increases as compared to the normal driving condition, as shown in Figure 2. As we can notice that the reconstruction error is low in the sunny driving scenario (i.e., because the driving conditions are similar to the training data). However, in adverse weather conditions such as snow, rain, and fog, the reconstruction error increases above the threshold (i.e., θ=0.05). In dense fog, the reconstruction error exceeds the threshold; hence, the car crashes on the side of the road. We observe that when the reconstruction error is increased above *0.05*, inconsistencies in the car behaviour increase (*i.e., the car no longer follows the center of lane and collision with roadsides*). Once the inconsistent behaviour is predicted, the safety guards are automatically activated to ensure safety in such scenarios. The designed safety guards in our approach are considered by following the safety standards for automated vehicles provided by the National Highway Traffic Safety Administration (NHTSA 2020) and ISO26262. In the following subsection, we explain the detail of our approach.

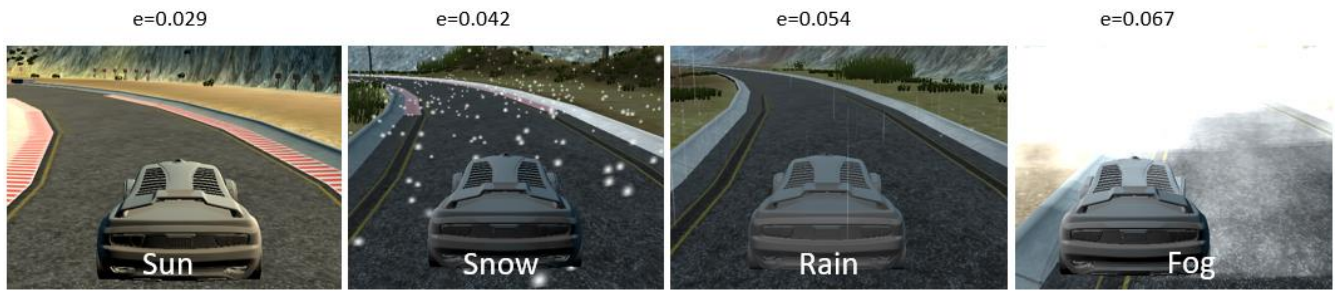

**Fig. 2** Different levels of reconstruction errors in response to different driving scenarios.

*A. Training of DeepGuard*

While training *DeepGuard*, we focus on behavioural cloning in which the ADS learns the lane-keeping behaviour of the human driver. The ADS is trained with images from camera sensors that record driving scenarios with steering angles from human drivers. The DNN model starts learning by discovering the features within the training data that represent the driving scenarios, such as the shape of the road and then predicting its corresponding steering angles, throttle, brakes, and speed. We train our inconsistent behaviour predictor with nominal data (*i.e., a sunny driving scenario).* Autoencoder-based anomaly detection systems are among the best techniques for detecting anomalies and are computationally very efficient. Figure 3 shows the steps involved in training the inconsistent behaviour predictor. The training of the inconsistent behaviour predictor consists of the following steps:

*1) Datasets:* We collected the dataset under nominal driving conditions. A dataset of 120,624 training images at 12 fps from all three tracks was obtained. Overall, we collected 34,160 images from Track 1 (Lake), 45,214 from Track 2 (Jungle), and 41,250 from Track 3 (Mountain). During the training, we recorded the training images with driving parameters such as speed, brake, and throttle. The speed of the car in the training mode was set at 30 m/h (i.e., mile/h).

2) *Autoencoders:* According to recent studies, 95% of traffic accidents are caused by human factors. Among these factors, abnormal driving is at the top of the list (Hu et al. 2017). Therefore, the need for ADSs has increased. However, an autonomous driving system itself sometimes exhibits anomalous behaviour when it faces unexpected driving conditions. Autoencoders are good at detecting anomalies, and recently they have been used in many domains, such as detecting anomalous driving scenarios in ADSs. There are three main components of autoencoders: the encoder, compressor (sometimes called latent space), and decoder. The autoencoder encodes the input image *x* using a function *f*, which is the latent space of the autoencoder. It then decodes the encoded value *f(x)* using another function *g(x)*, which is the decoder of the autoencoder, to create an output value that is the same as the input value.

The difference between the input vector *x* and the generated output vector $\bar{x}$ is called the reconstruction error. The objective of the autoencoder is to minimize the reconstruction error such that the input vector and output vector remain the same. The loss function $L(x, \bar{x})$ actually measures the difference between the input value and its low-dimensional reconstructed output. The mean square error *(MSE)* is a widely used loss function in autoencoders. The *MSE* for the autoencoders used in our approach can be obtained using Equation 1.

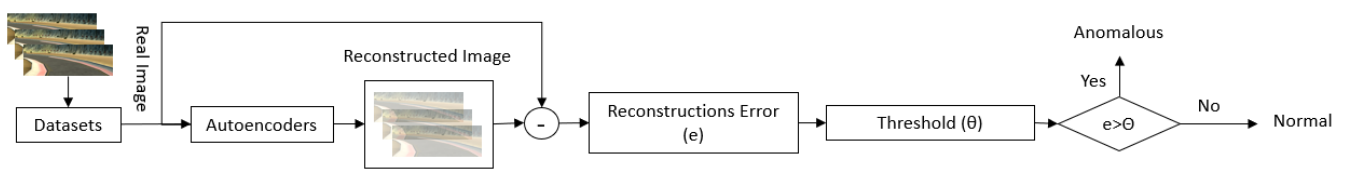

**Fig. 3** Training of autoencoder-based inconsistent behaviour predictor using normal driving dataset.

$$L(x, \bar{x}) = \frac{1}{m} \sum_{i=1}^{i=m} (x_i - \bar{x}_i)^2 \qquad (1)$$

Where *m* is the output dimension of the image, *and* $x_i$ and $\bar{x}_i$ are the elements of $x$ and $\bar{x}$ respectively. Since the loss function measures the difference between the input value and output value, we can use the MSE of autoencoders as a behavioural monitor at runtime. For example, we can use the increasing trends in MSE to warn the ADS under unexpected driving conditions. In unexpected driving conditions, the MSE will be higher than the driving conditions in which the ADS is trained, and we can effectively manage the inconsistent behaviour of the ADS based on the reconstruction error. The autoencoders are trained by unsupervised learning, which reconstructs the input image by maintaining a low reconstruction error. It consists of an encoder and decoder. A single hidden layered autoencoder has an encoder and decoder, as given in Equations 2, and 3 respectively (An and Cho 2015). The reconstruction error is given by Equation 4.

$$\mathbf{h} = \sigma(Wxh^x + b_{xh}) \qquad (2)$$

$$\mathbf{z} = \sigma(Whx^h + b_{hx}) \qquad (3)$$

$$Reconstruction\ Error = ||x - z|| \qquad (4)$$

Equation (2) represents the encoding of the input image in which the input image *x* is mapped with the hidden representation *h*, while equation (3) represents the decoding of the input image from its hidden representation. Equation (3) maps the hidden representation *h* back to its original input image. However, the reconstruction error is the difference between the input image *x* and the reconstructed image *z*. The reconstruction error of a single hidden layer autoencoder can be calculated using Equation (4).

*3) Reconstruction Error:* We used multilayered autoencoders in our approach. In this case, the pixel-wise reconstruction error can be calculated using the following equation proposed by the authors in (Stocco et al. 2020) given in equation 5.

$$d(x, x') = 1/WHC \sum_{i=1, j=1, c=1}^{W, H, C} (x[c][i,j] - x'[c][i,j])^2 \qquad (5)$$

In the above equation, the reconstruction error $e = d(x, x')$ is calculated by comparing the individual pixels of the input image *x* and the reconstructed image $x'$ and taking the mean pixel-wise squared error. *W* is the width, *H* is the height, and *C* is the channel of the image.

*4) Threshold Estimation:* To estimate the threshold, we take the inverse of the cumulative distribution of the gamma fitting. For example, we have a set of reconstruction errors and consider the false alarm rate to be 0.02. In this case, we can estimate the threshold θ by taking the inverse of the cumulative distribution of the Gamma parameters (Stocco et al. 2020) using Equation 6.

$$F(x): \theta = F^{-1}(1 - false\ alarm\ rate) \qquad (6)$$

---
**Algorithm 1. Training of Autoencoders**

---

**Input:** dataset (Nominal Data) $x_1, x_2, x_2, ...., x_n$

**Output:** encoder $f\varphi$ and decoder $f\varphi'$

1. $\varphi, \varphi' \leftarrow$ parameters initialization
2. **do**
3. $$e = \sum_{i=1}^{N} ||x_i - f\varphi'(f\varphi(x_i))||$$
4. $\varphi, \varphi' \leftarrow$ Using the gradient of $e$ update parameters
5. **While** the convergence of $\varphi, \varphi'$

The estimated $\theta$ is then used to differentiate between the normal driving conditions (i.e., $e<\theta$) and anomalous driving conditions (i.e., $e>\theta$). During simulations, we tested many threshold $\theta$ corresponding to the false alarm rate above *0.05*, that is, *0.06,* and below *0.05*, that is, 0.01. However, we found that the threshold corresponding to the false alarm rate of 0.01 makes the prediction of inconsistent behaviour of ADSs useless as the true positive rate (TPR) and false-positive rate (FPR) becomes zero or equivalent to zero. Our objective was to increase the TPR and decrease the FPR. We observed a lower FPR and higher TPR when setting a threshold corresponding to the false rate of *0.05*.

Algorithm 1 represents the generalized algorithm for training autoencoder-based inconsistent behaviour predictors. We use the best-performing autoencoders, such as the variational autoencoder, deep autoencoder, and denoising autoencoder as inconsistent behaviour predictors.

In the following subsection, we explain the second part of our approach, which is the prevention of inconsistent behaviour in ADS by executing safety guards.

*B. Prevention of Inconsistencies in ADS Behaviour*

DNN-based autonomous systems are safety-critical; therefore, these systems must endure runtime faults and preserve safety when the system faces inconsistencies in the DNN model. Typical methods for safety assurance, such as testing and validation, do not meet the requirements of the fail-operational and safe operational concepts. Therefore, we use the concept of runtime execution of safety guards to improve the safety of ADSs. The inconsistent behaviour prevention system ensures the safety of the ADS through the synthesis of runtime execution of safety guards.

The camera mounted on the ADS captures the driving scenarios continuously, and the autoencoder in the inconsistent behaviour predictor recreates the driving scenarios. Then, the reconstruction error patterns are feed to the time series analyzer. We use the autoregressive time-series analysis model to predict potential inconsistent behaviour. The increasing trends in the reconstruction error of the autoencoders indicate that the input images are anomalous. After comparing the filtered reconstruction error with the hardcoded threshold $\theta$, the inconsistent behaviour predictor determines whether the driving scenarios are anomalous or normal. Based on the different levels of reconstruction error, different levels of safety guards are applied to ensure the safe driving of ADSs.

Time-series analyses are used to enhance the predictability of the predictive models. The time-series model uses previous input/output values to predict future values. It can be applied to the reconstruction error produced by autoencoders to identify trends and predict driving behaviour. Different types of time-series analysis models are available to predict future values based on past values, such as vector autoregression, autoregressive, integrated, and moving average models. These models are widely used time series analysis models in many domains to predict future values based on past values to perform useful operations. However, in our approach, we use a simple autoregressive time series analysis model to predict the potential inconsistent behaviour caused by unexpected driving conditions.

A simple model of the autoregressive time-series analysis model used in our approach is given in Equation (7). The autoregressive model predicts the desired variable using a linear combination of past variable values. An autoregressive of order $m$ is given as follows: Variable $x_t$ can be predicted using past values such as $x_{t-1}, \ldots, x_{t-m}$.

$$x_t = c + \phi_1 x_{t-1} + \phi_2 x_{t-2} + \cdots + \phi_m x_{t-m} + E_t \quad (7)$$

Where the $E$ represents the error terms or sometimes called white noise.

We want to ensure that the predefined set of safety guards is always satisfied by the system, even in the presence of an unknown failure. We used the safety guard concept introduced by the authors in (Wu et al. 2017). These safety guards are activated similar to *runtime enforcers* at runtime when an anomaly in the image stream is predicted.

Let us consider that the ADS takes input $I$ through camera sensors and produces the driving parameters as output ($O$). The input $I$ is defined as follows:

$$I = x_1, x_2, x_3 \ldots x_i$$

Where $x_1, x_2, x_3 \ldots x_i$ are the input images.

Suppose that the $SR_1$ and $SR_2$ are the safety requirements for ADSs that are defined as:

$$SR_1 = The\ vehicle\ must\ follow\ the\ center\ of\ the\ lane$$
$$SR_2 = No\ collision$$

Given that the safety requirement is denoted ($\psi$) and defined as follows:

$$\psi\ (SR_1,\ SR_2)$$

However, in some conditions, such as unexpected driving conditions, the ADS sometimes violates the safety requirements. This means that in anomalous driving conditions such as *fog, rain, and snowy* environments, the ADS may violate safety requirements $\psi$ by making *collisions with roadsides* or *the vehicle no longer follows the center of the lane*. In such scenarios, our approach automatically activates a predefined set of safety guards ($SGs$). $SGs$ are activated in response to different levels of violations of $\psi$.

Let us consider that $L_1$, $L_2$, and $L_3$ are the different levels of safety guards for different levels of violations of $\psi$. In this case, the $SGs$ can be defined as follows:

$$Safety\ gaurds = SGs(L_1, L_2, L_3\ )$$

Where:

$$L_1 = Leve\ 1\ safety\ guards$$
$$L_2 = Leve\ 2\ safety\ guards$$
$$L_3 = Leve\ 3\ safety\ guards$$

The level of violation of $\psi$ is determined by the value of the reconstruction error (i.e., $e$) produced by the autoencoders for each input image $x_i$. Thus, whenever the ADS violates the $\psi$, *DeepGuard* automatically activates a set of safety guards based on the level of violation of $\psi$.

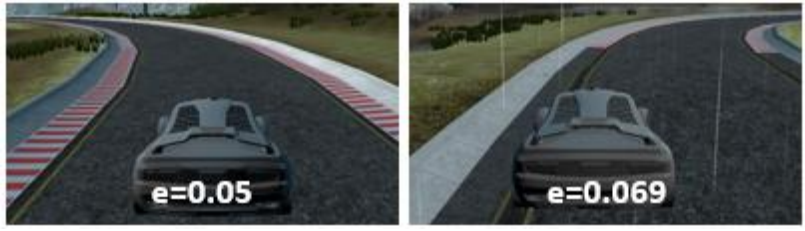

**Fig. 4** The ADS still produces correct driving parameters after applying a self-healing safety guard (i.e., image with *e=0.05*). However, when the reconstruction error *e* reaches up to *0.069*, the vehicle no longer follower the center of the road.

*DeepGuard* takes the original input *I* of the ADS and produces a reconstruction error ($e$). (Fig. 1). For each input image $x_1, x_2, x_3 \ldots x_i$, the autoencoders reconstruct the images $\acute{x}_1, \acute{x}_2, \acute{x}_3 \ldots \acute{x}_i$ and measure reconstruction errors $e = e_1, e_2, e_3 \ldots, e_i$ respectively. Hence, *DeepGuard* predicts the potential violation of $\psi$ based on $e$ and automatically activates a set of *SGs* by producing its own output $O'$. Thus, $O'$ (i.e., *DeepGuard* output) overrides the $O$ (i.e., ADS output) when it violates the safety requirements $\psi$ in unexpected driving contexts. Our goal is that the $O'$ always activates safety guards for the ADS in unexpected

| | Algorithm 2. Autoencoder and time series analysis-based inconsistent behaviour prevention algorithm |
|---|---|
| | **Input:** dataset (Nominal Data) *X, Anomalous dataset* $x_i$ i=1,2,3…, N, Threshold θ |
| | **Output:** Filtered reconstruction error $e = \|x - \bar{x}\|$ |
| 1 | φ, φ' ← Train autoencoders with nominal data X |
| 2 | **For** i=1 && i<=N **do** |
| 3 | *reconstruction error (i)* $$e_{(i)} = \|x_i - f\varphi'(f\ \varphi(x_i))\|$$ |
| 4 | Apply Autoregressive for $e_{(i)}$ to predict inconsistencies in ADS behaviour |
| 5 | **If** $e_{(i)}$> θ **then** |
| 6 | *$x_{(i)}$* is **anomalous driving context** |
| 7 | **If** $e_{(i)}$> θ && ($e_{(i)}$>=0.05&& $e_{(i)}$<0.059) |
| 8 | *activate level 1 safety guards* |
| 9 | **else If** $e_{(i)}$> θ &&($e_{(i)}$>=0.059 && $e_{(i)}$<0.069) |
| 10 | *activate level 2 safety guards* |
| 11 | **else if** $e_{(i)}$> θ && $e_{(i)}$>=0.069 |
| 12 | *activate level 3 safety guards* |
| 13 | **Else** |
| 14 | *activate level 3 safety guards* |
| 15 | **end if** |
| 16 | **Else** |
| 17 | *$x_{(i)}$* is normal driving context |
| 18 | **end if** |
| 19 | **end for** |

driving conditions. However, the $O'$ must not override the original output $O$ when the driving conditions are normal. The *DeepGuard* output $O'$ interferes with the ADS output $O$ only under anomalous driving conditions. The definitions of the above-mentioned safety guards are given below.

*1) Level 1 Safety Guards:* The first level of safety guards consists of the following safety measures to prevent inconsistencies in ADS behaviour before it occurs.
*Self-healing:* This safety guard is applied when the trends in increasing reconstruction error remain for a very short time, and the reconstruction error value is less than 0.055. The ADS with this safety guard heals itself when the anomalous driving scenario is for a short period. After the anomalous driving scenario passes the ADS, drives normally.
*Lane Departure Warning:* When the self-healing safety guard could not control the potential inconsistencies in ADS behaviour and the increasing trends in reconstruction error exceed *0.055*, the lane departure warning is triggered to alert the human driver. Remember that our safety requirement is to keep the vehicle in the center of the lane and avoid collisions. Therefore, if the reconstruction error continues to increase, we assume that the vehicle is no longer following the center of the road; hence, we want to trigger the lane departure warning to the human driver so that the human driver can take over the control and bring the vehicle to the safe zone. Figure 4 shows the effect of the increase in reconstruction error. Usually, when the reconstruction error *e* remains in the range of *0.05-0.059,* we observe that the first level of safety guards is sufficient to prevent potential inconsistent behaviour.

*2) Level 2 Safety Guards:* This level of safety guards is applied for reconstruction errors ranging from *0.059-0.069*. *Level 2 safety guards* include *self-healing*, *lane departure warning*, *and speed reduction*.
*Reduction Speed:* One of the most influential factors in a traffic crash is overspeeding of vehicles. Reducing speed in dangerous situations can stop crashes. Therefore, we applied this technique to prevent crashes using our approach. The idea is to reduce the autonomous car speed and apply an emergency brake to stop the vehicle once the inconsistencies in ADS behaviour are predicted.

*3) Level 3 Safety Guards:* Most of the collisions were observed during our experiment when e>0.069. Therefore, with the combination of Level 1 and Level 2 safety guards, and some extra safety guards (i.e., automatic emergency brake and autonomous mode disengagement), we designed Level 3, safety guards.
*Automatic Emergency Brake:* The automatic emergency brake is applied with a combination of other safety guards, such as a reduction in vehicle speed. After applying *Level 1 and Level 2 safety guards*, if the ADS misbehaves, then *DeepGuard* activates the *automatic emergency brake* to stop the vehicle from collisions.
*Safe Disengagement of Autonomous mode:* An autonomous mode disengagement safety guard is designed by following the level 4 autonomous car safety feature provided by NHTSA in which the car drives autonomously; however, the human driver must sit on the driving seat in order to respond to any emergency detected by the ADS. This safety guard is activated after the automatic emergence brake stops the vehicle so that the ADS does not go for self-healing and starts driving again. After the automatic emergence brake stops the vehicle, the control of the ADS hands over to the human driver so that the driver brings out the ADS from the hazardous driving environment. Once the car reaches the safe zone, it starts driving by itself again.

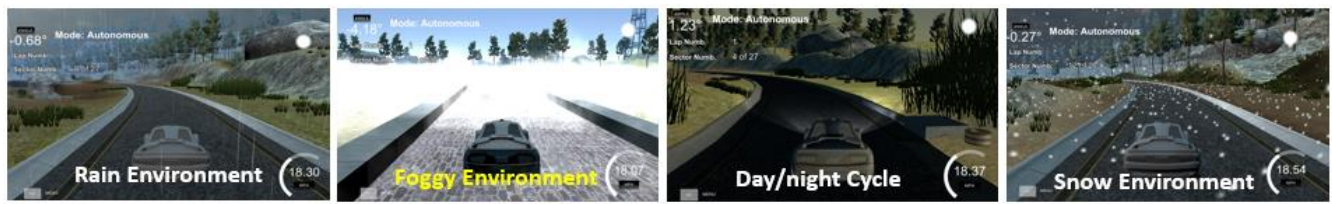

**Fig. 5** Different unexpected driving conditions such as foggy, day/night cycle, raining, and snow conditions.

The inconsistent behaviour predictor uses reconstruction error as an anomaly score. Data points with high reconstruction errors are considered as anomalous driving scenarios. Algorithm 2 represents the algorithm for the inconsistent behaviour prevention system of our approach.

## IV. Evaluation

### A. DNN-Based Autonomous Car Models

To evaluate the proposed approach, we deploy *DeepGuard* on three existing DNN-based ADSs: Chauffeur (Team Chauffeur 2016), Epoch (Epoch Team 2016), and Nvidia's DAVE-2. Owing to the robustness of these models, we chose these DNN models to evaluate *DeepGuard*. Another reason for choosing these models is that they are open-source, and we can easily evaluate them using the simulator. The chauffeur model uses a convolutional neural network to extract the driving features from the input image, and another recurrent neural network is used to predict the steering angle from previous consecutively extracted driving features. The Epoch ADS model consists of only a three-layered CNN model, whereas the Nvidia DAVE-2 consists of three convolutional layers with five fully connected layers.

### B. Simulation Platform

We used a simulation environment to evaluate the proposed approach. The reason for using the simulation environment is that the testing of ADS inconsistent behaviour, such as collision, requires the creation of unexpected driving conditions and requires a way to record these crashes during testing. Thus, the simulation environment provides a way to create unexpected driving conditions for testing and evaluating the ADS. The simulator-generated data yielded similar predictions to those generated by real-world datasets, as mentioned in (Haq et al. 2020). Thus, we used the modified version of the Udacity (Udacity 2017) simulator provided by *SELFORACLE*. The Udacity simulator is built on (Unity Technologies 2021) which is a popular game engine. The modified version of the Udacity simulator provided by *SELFORACLE* has two modes: the first is the *training mode,* in which the user can collect the driving datasets by manually controlling the car, while in the second mode, that is, *autonomous mode*, in which car is controlled by DNN-based ADS models. This simulator has an *unexpected driving context generator* and *a collision detection system.* Hence, we chose to use this simulator to evaluate *DeepGuard*.

*1) Unexpected driving condition generator:* The *SELFORACLE* simulator unexpected driving context generator creates unexpected driving conditions in autonomous mode as compared to the driving conditions considered in the training mode (i.e., *nominal condition*). Two driving conditions were considered as unexpected driving conditions. In the first condition, the *SELFORACLE* simulator gradually changes the lightning of the conditions on the track to simulate the day/night cycle.

The second condition is related to weather effects. In this condition, raining, snowing, and foggy effects are created in the simulator as anomalous driving conditions to evaluate the inconsistent behaviour predictor. To create the snow, rain, and foggy effect, the particle system (Unity 2017) is provided by Unity. This system can simulate the physics of a cluster of particles. The particle system spawns predefined clusters of particles to create different effects, such as fog, rain, and snow. For example, the rain particles emission rate ranges from 100 (light rain) to 10,000 particles/s (heavy rain); to create foggy conditions the number of particles ranges from 100 to 2,000 particles/s, and the snow effects can be created using 100 to 800 particles/s. Figure 5 shows the different unexpected driving conditions created to evaluate the effectiveness of the approach during autonomous mode after training the ADS with nominal data.

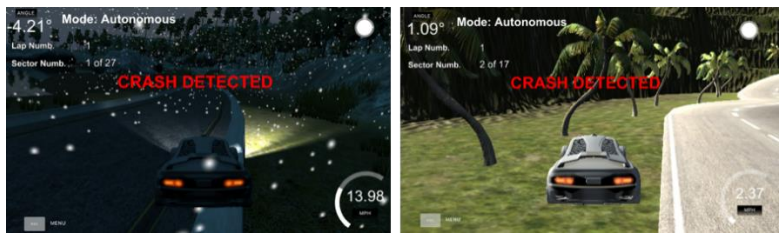

**Fig. 6** Crash Detector detects that the ADS crashes on different tracks.

*2) Collision Detector:* The definition of safety-critical inconsistent behaviour in our approach is that the car collides *with the roadsides*, and the second one is that the *car no longer follows the center of the lane*. Therefore, we use the automated collision/out-of-bound episode system detection system of *SELFORACLE* to record the *collisions* and *lane departure* during the occurrence of unexpected driving conditions. The automated out-of-bound episode detection system is based on colliders and simulates the physical interaction between objects. When the car is on the road, it means the car is on the actual track; however, when the car collides with any other object in the scene, the collider callback registers that it has crashed with some object. The automated restart mechanism resumes the ADS from a safe position after crashes. This technique allowed us to record various numbers of simulations without manually restarting the simulator. Figure 6 illustrates how the crash detection system works.

*C. Experimental Method:*

In the following subsection, the detail of experimental method and simulations setup is discussed in detail.

1) Training Data Generation:

Table 1 presents the details of data collection during the training mode. The speed was set to 30 miles/hr to capture the driving behaviour, such as lane centering and no collision. The difference in the number of images is due to the length of the tracks. These training images were collected at a rate of 12 frames/s.

Table 1. Training Dataset from all three tracks

| Tracks | Number of Images | Orientation |
|---|---|---|
| Track 1 (Lake) | 34,160 | Forward/Reverse |
| Track 2 (Jungle) | 45,214 | Forward/Reverse |
| Track 3 (Mountain) | 41,250 | Forward/Reverse |
| Total Number of Images= 120,624 | | |

*2) Training of ADS Models:* We considered three (i.e., Chauffeurs, Nvidia DAVE-2, and EPOCH) ADS models to evaluate the approach as mentioned earlier. Therefore, we trained all these ADS models using the collected training datasets. We used *120,624* images to train the model. The details of the training set of the ADS models are given in Table 2.

Table 2. Training setup and system specification detail.

| | Total Number of Images | 120,624 |
|---|---|---|
| Data | Data Augmentation Techniques | Flipping, translations, and brightness, etc. |
| | Color | RGB |
| | Image Size | 70 x 210 |
| Training | Number of Epoch | 500 |
| | Batch size | 256 |
| System Specification | Processor | i7 |
| | Memory RAM | 32 GB |
| GPU Detail | GPU Model | GeForce RTX 2060 |
| | GPU Dedicated Memory | 6 GB |

*3) Evaluation Dataset:* To generate the evaluation data, we collected data from all three tracks in eight different conditions (e.g., rain, fog, day/night cycle, rain at night, etc.). We first executed a simulation similar to the training set without injecting the anomalous driving conditions to record the false alarm rate. Then, we executed the simulations by injecting anomalous conditions such as rain, snow, fog, and day/night cycles. The details of the evaluation dataset are given in Table 3.

Table 3. Evaluation dataset with known and unknown conditions.

| Tracks | Number of Images | Conditions |
|---|---|---|
| Track 1 (Lake) | 122,773 | Unknown |
|  | 31,250 | Known |
| Track 2 (Jungle) | 221021 | Unknown |
|  | 47,521 | Known |
| Track 3 (Mountain) | 187,048 | Unknown |
|  | 41,253 | Known |
| Total Number of Images= 650,866 ||| 

We used *SELFORACLE* as a baseline for our approach. However, the healing system in their approach was followed by *misbehaviour windows*. This means that the healing system is only activated after misbehaviour has occurred. In contrast to the *SELFORACLE* approach, our healing system, that is, the safety guards approach is activated before the inconsistent behaviour occurs.

*D. Results*

We use the following metrics to analyze the results of *DeepGuard:*

*True Positive (TP):*

*TP* is defined when the *DeepGuard* predicts the potential inconsistent behaviour and activates safety guards successfully. When the reconstruction errors increase above the estimated threshold, *DeepGuard* triggers alarms and activates safety mechanisms.

*False Negatives (FN):*

False-negative is defined when the *DeepGuard* does not predict potential inconsistent behaviour and fails to activate safety guards during anomalous driving conditions.

*False Positives (FP):*

A false positive is defined when *DeepGuard* triggers a warning for normal driving conditions. For example, the driving condition is normal (i.e., sunny conditions) and DeepGuard triggers alarms and activates unnecessary safety guards, then such scenarios are defined as false positives.

*True Negatives (TN):* A true negative is defined when *DeepGuard* predicts normal driving conditions and triggers no alarm, nor any safety guards are activated. In this condition, the ADS drives normally, and there is no role of DeepGuard in terms of safeguarding the ADS in normal driving scenes. Because the driving scenarios are normal, *DeepGuard* detects them as normal. Figure 7 illustrates the metrics used by *DeepGuard*.

|  | Actual (Anomalous) | Actual (Normal) |
|---|---|---|
| Predicted (Anomalous) | True Positives (Triggers Alarm & Activates Safety Guards) | False Positives (False Alarm & Unnecessary Activation of Safety Guards) |
| Predicted (Normal) | False Negatives (No Alarm & No Safety Guards are Activated) | True Negatives (Correct Detection of Normality. Hence, no Alarm & no Safety guards are Activated) |

**Fig. 7** Metrics used to evaluate the *DeepGuard*.

Our objective is to maximize the TPR while minimizing the FPR.

$$TPR = \frac{TP}{TP + FN} \tag{8}$$

$$FPR = \frac{FP}{TN + FP} \tag{9}$$

The TPR and FPR can be obtained using Equations 8 and 9, respectively. A higher TPR indicates a higher true alarm (i.e., the correct prediction of anomalous driving conditions), and
A lower FPR means a lower false alarm, that is, predicting the safe driving scenario as unsafe, which results in triggering a false alarm and activating unnecessary safety guards.

We also include another matrix to evaluate the results of *DeepGuard*, such as the area under the precision-recall curve (AUC-PRC). We did not consider the area under the curve of the receiver operating characteristics (AUC-ROC) as it is inefficient to score the results when the data are highly imbalanced as mentioned in (Davis and Goadrich 2006). In our case, as compared to the normal driving images, the anomalous driving conditions are very rare; hence, using the AUC-ROC matrix is not useful.

In order to evaluate the proposed approach, we define the following questions.

**RQ1 (Performance):** How good is the performance of *DeepGuard* in terms of predicting the inconsistent behaviour of ADSs? What are the best-performing autoencoders to predict the inconsistent behaviour of ADSs?

**RQ2 (Effectiveness of Safety Guards):** How effective are safety guards to ensure the safety of the ADS when inconsistent behaviour is predicted?

**RQ3 (Comparative Analysis):** How *DeepGuard* is better than *SELFORACLE's* autoencoder-based misbehaviour prediction and *DeepRoad's* generative adversarial network and online input validation technique to predict inconsistent behaviour of the ADS?

In the following, we answer the above-mentioned research questions.

**Performance (RQ1).** In response to RQ1, Table 4 presents the performance of *DeepGuard* for each ADS model. Column 4 presents the AUC-PRE measure, which is independent of thresholds. Columns 5 to 12 represent the performance of each variant of *DeepGuard* when the threshold was set corresponding to a false alarm rate *of 0.05*. We considered only the threshold corresponding to a false alarm rate of *0.05*. When the value of the false alarm rate is lower than *0.05,* FPR and TPR become zero, making the prediction useless. On each ADS model, VAE and DAE performed outstandingly on AUC-PRE and other threshold-

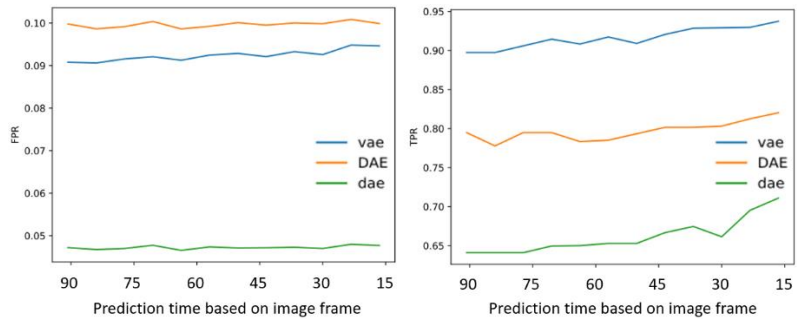

**Fig. 8** TPR and FPR on Chauffeur model under anomalous conditions.

Table 4. Comparative analysis of simulation results and healing systems of *DeepGuard* VS *SELFORACLE* and *DeepRoad*.

| Approach | ADS Model | AE | AUC-PRC | $\theta$ = (1-false alarm) =1-0.05=0.95 | | | | | | | | Healing Strategies | [1]Cond. |
|---|---|---|---|---|---|---|---|---|---|---|---|---|---|
| | | | | TP | FP | TN | FN | TPR | FPR | F1 | Prec. | | |
| SELFORACLE and DeepRoad | DAVE-2 | VAE | 0.354 | 149 | 304 | 1,970 | 47 | 0.760 | 0.134 | 0.459 | 0.329 | Self-healing system | After inconsistent behaviour has happened. |
| | | [2]dae | 0.330 | 104 | 210 | 2,679 | 92 | 0.531 | 0.073 | 0.408 | 0.331 | | |
| | | SAE | 0.336 | 138 | 260 | 1,877 | 58 | 0.704 | 0.122 | 0.465 | 0.347 | | |
| | | CAE | 0.290 | 6 | 22 | 4,208 | 190 | 0.031 | 0.005 | 0.054 | 0.214 | | |
| | | LSTM | 0.357 | 16 | 34 | 3,990 | 177 | 0.083 | 0.008 | 0.132 | 0.320 | | |
| | | DeepRoad | 0.198 | 65 | 344 | 3,170 | 131 | 0.332 | 0.098 | 0.215 | 0.159 | | |
| | Epoch | VAE | 0.391 | 158 | 331 | 1,952 | 51 | 0.756 | 0.145 | 0.453 | 0.323 | | |
| | | dae | 0.399 | 112 | 188 | 2,720 | 97 | 0.536 | 0.065 | 0.440 | 0.373 | | |
| | | SAE | 0.386 | 147 | 284 | 2,026 | 62 | 0.703 | 0.123 | 0.459 | 0.341 | | |
| | | CAE | 0.310 | 6 | 23 | 3,661 | 203 | 0.029 | 0.006 | 0.050 | 0.207 | | |
| | | LSTM | 0.385 | 23 | 34 | 3,503 | 175 | 0.116 | 0.010 | 0.180 | 0.404 | | |
| | | DeepRoad | 0.213 | 70 | 308 | 2,917 | 139 | 0.335 | 0.096 | 0.239 | 0.185 | | |
| | Chauffeur | VAE | 0.242 | 98 | 392 | 3,700 | 23 | 0.810 | 0.096 | 0.321 | 0.200 | | |
| | | dae | 0.203 | 78 | 281 | 5,045 | 43 | 0.645 | 0.053 | 0.325 | 0.217 | | |
| | | SAE | 0.241 | 96 | 354 | 3,650 | 25 | 0.793 | 0.088 | 0.336 | 0.213 | | |
| | | CAE | 0.172 | 7 | 34 | 8,127 | 114 | 0.058 | 0.004 | 0.086 | 0.171 | | |
| | | LSTM | 0.240 | 11 | 41 | 7,879 | 111 | 0.090 | 0.005 | 0.126 | 0.212 | | |
| | | DeepRoad | 0.098 | 37 | 594 | 5,564 | 84 | 0.306 | 0.097 | 0.098 | 0.059 | | |
| DeepGuard | Epoch | VAE | 0.425 ↑ | 207 | 316 | 2,051 | 52 | 0.800 ↑ | 0.134 | 0.530 | 0.396 | Safety guards approach 1. Level 1 2. Level 2 3. Level 3 | Before inconsistent behaviour has happened |
| | | [3]DAE | 0.410 ↑ | 194 | 334 | 2,093 | 65 | 0.750 ↑ | 0.138 | 0.493 | 0.367 | | |
| | | dae | 0.418 ↑ | 166 | 222 | 2,945 | 92 | 0.641 | 0.070 ↓ | 0.513 | 0.428 | | |
| | DAVE-2 | VAE | 0.356 | 184 | 309 | 1,796 | 36 | 0.837 ↑ | 0.147 | 0.516 | 0.373 | | |
| | | DAE | 0.355 | 177 | 304 | 1,923 | 43 | 0.805 ↑ | 0.136 | 0.505 | 0.368 | | |
| | | dae | 0.352 ↑ | 132 | 221 | 3057 | 88 | 0.602 | 0.067 ↓ | 0.461 | 0.375 | | |
| | Chauffeur | VAE | 0.245 ↑ | 120 | 338 | 3,241 | 8 | 0.937 ↑ | 0.099 | 0.410 | 0.262 | | |
| | | DAE | 0.264 ↑ | 105 | 400 | 3,606 | 23 | 0.821 ↑ | 0.100 | 0.332 | 0.207 | | |
| | | dae | 0.238 ↑ | 90 | 278 | 5,555 | 37 | 0.713 | 0.047 ↓ | 0.366 | 0.264 | | |

[1]Cond. This column includes the conditions (before/after) for healing system.
VAE: Variational Autoencoder, [2]dae: Deep Autoencoder, SAE: Simple Autoencoder, CAE: Convolutional autoencoder, LSTM: Long Short-Term Memory, and [3]DAE: Deniosing Autonecoder dependent metrics such as TPR and FPR. We did not consider the AUC-ROC as it does not provide better information for analysis when the data are heavily imbalanced (Saito and Rehmsmeier 2015). Compared to the normal images, there are very few anomalous images in our datasets. Hence, we do not consider AUC-ROC.

On the TPR and FPR metrics, VAE and DAE are the best among all variants. Overall, the performances of VAE andDAE are 80% vs. 13 % (TPR vs. FPR of VAE), and 75% vs. 13 % (TPR vs. FPR of DAE) on the Epoch model. The TPR and FPR of VAE were 83% and 14%, respectively, while the TPR and FPR of DAE were 80% and 13 %, respectively, On the DAVE-2 model. The performance of VAE and DAE was observed at its best on the Chauffeur ADS model, which is 93% vs. 9% (TPR vs. FPR of VAE), and 82% vs. 10% (TPR vs. FPR of DAE). Figure 8 shows the TPR and FPR of different variants of *DeepGuard*.

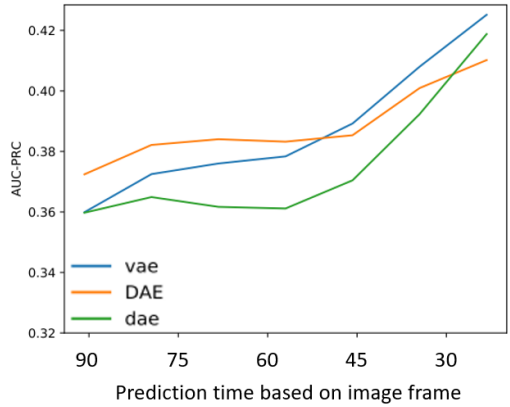
*Fig. 9* AUR-PRE for VAE, DAE and dae on DAVE-2 model.

The deep autoencoder (i.e., dae) could not perform well compared with *SELFORACLE*. However, it still outperformed the *DeepRoad* approach. The performance of the *DeepGuard* variants is more than three times higher than that of *DeepRoad*. The false-positive rate was higher than the set threshold value *(*corresponding to false rate = *0.05).* The FPR is measured for tracks (Udacity's three tracks) with injected anomalies. However, in those conditions when the intensities of the injected anomalies are low, these conditions are not critical to yield any inconsistent behaviour in ADSs. It was also expected that the false alarm rate may be higher in anomalous driving conditions than in normal driving conditions. Hence, the FPR is over-approximated in an anomalous driving context. Answering the second part of RQ1, VAE and DAE performed the best among all variants of the *DeepGuard*. The TPR is significantly higher with VAE and DAE (80% and 75%, respectively). On DAVE-2 TPR for VAE and DAE were 83% and 80%, respectively, and for the Chauffeur model, the TPR was 93% and 82% for VAE and DAE, respectively. The performance of the *DeepGuard* variant on AUR-PRE is shown in Figure 9. From figure 9, it is clear that it is difficult to predict anomalous inconsistent behaviour when the ADS is far from the anomalous region. It can be seen that the performance continues to rise and never drops. However, the closer to the anomalous region, the better the prediction of inconsistent behaviour. Conversely, the more the closer to the anomalous region, the lesser the reaction time to activate safety guards.

**Effectiveness of safety guards (RQ2).** In contrast to *the SELFORACLE* and *DeepRoad* approaches, our healing system consists of a set of predefined safety properties that are always satisfied by the ADS. These safety guards have shown excellent performance during the simulations. Our simulation results showed that the predicted inconsistent behaviours were successfully controlled by designed safety guards, and some of the minor inconsistencies that last for a short period of time were mitigated by the self-healing system. *DeepGuard* activates safety guards to prevent the predicted inconsistent behaviour in advance, regardless of the defined reaction time. In contrast to the *SELFORACLE* approach, the *DeepGuard* activates the safety guards without going through a fixed reaction time. In the *SELFORACLE* approach, the reaction time is fixed (i.e., *4* seconds). However, when any new anomalous context is detected within the reaction time, the *SELFORACLE* cannot anticipate the potential inconsistencies in ADS behaviour for newly predicted anomalous context resulting in low prediction (i.e. 77%) as well as more crashes. Hence, in our approach, the healing system is activated just after the detection of anomalous context that helps to activate safety guards and a higher predictability rate.

Table 5. Effectiveness of the healing system of *DeepGuard* in preventing the inconsistent behaviour of ADSs.

| Models | Autoencoders | Prediction of injected anomalies (%) | Preventions of inconsistent behaviour by safety guards (%) | Total |
|---|---|---|---|---|
| Chauffeur | VAE | 93 | 91 | |
| | DAE | 82 | 89 | 89% |
| | dae | 71 | 87 | |
| DAVE-2 | VAE | 83 | 88 | |
| | DAE | 80 | 89 | 87% |
| | dae | 60 | 86 | |
| Epoch | VAE | 80 | 89 | |
| | DAE | 75 | 87 | 87% |
| | dae | 64 | 86 | |

Overall, these designed safeguards prevent 89% of the predicted inconsistent behaviours in the Chauffeur ADS model. When the ADS models are DAVE-2 and Epoch, these safety guards prevent 87% of all predicted inconsistent behaviours. Prevention of inconsistent behaviour means that it prevents the ADS from violating the safety specification, which is not to make *collisions* and *follow the center of the lane* while driving. Table 5 illustrates the efficacy of the healing system of our approach.

**Comparative Analysis (RQ3).** In response to RQ3, our approach outperforms DeepRoad and SELFORACLE in terms of the predictability of inconsistent behaviour. Column 4 in Table 4 shows the AUC-PRE matrices of the VAE, DAE, and dae. The AUR-PRE column in Table 4 shows that the *DeepGuard* approach is superior to both *SELFORACLE* and *DeepRoad*. In terms of the TPR rate, our best-performing variant is VAE, which can predict 93% as compared to *SELFORACLE*'s best-performing variant VAE, which can only predict 77% of inconsistencies in ADS behaviour. The prediction rate of *DeepRoad* in terms of the TPR was 32%. The *DeepGuard* approach predictability rate is three times higher than that of *DeepRoad*.

Our safety-guards-based healing systems can prevent up to 89% of all predicted inconsistencies in ADS behaviour. Compared to *the SELFORACLE* and *DeepRoad* approaches, their healing systems only consist of a self-healing system. In terms of activation of the healing system, *DeepRoad* and *SELFORACLE*'s healing systems are activated after inconsistencies in driving behaviour have occurred. This is unrealistic in the real world. More specifically, the healing systems of *SELFORACLE* always follow the *misbehaviour windows*, which means that the healing system will be activated after misbehaviour occurs. We argue that once inconsistencies in driving behaviour occur, activating the healing system becomes useless. This is because the safety of the ADS is already compromised due to misbehaviour. Therefore, in contrast to *SELFORACLE* and *DeepRoad*, our approach activates different levels of safety guards for different levels of inconsistencies in ADS behaviour before it violates the safety properties. Table 4 shows a comparative analysis of our approach with the *SELFORACLE* and DeepRoad approaches.

*E. Limitations of DeepGuard*

The first limitation of our approach is that we generated unexpected driving conditions in a customized manner in the simulation environment. In the real world, there may be hundreds of thousands of unexpected driving conditions. However, this limitation can be handled by designed safety guards, as the predefined safety guards are always satisfied by the systems, even in the presence of unknown faults such as unexpected driving conditions. Another reason for creating unexpected driving conditions in a customized way is that our approach requires no knowledge about the ADS model; hence, the *DeepGuard* model can be attached to any ADS model. Any unexpected driving conditions faced by the ADS can be predicted by the *DeepGuard* and ensuring safety by executing the predefined safety guards. The performance of the *DeepGuard* may vary depending on the choice of ADS models and training data. Poor training of the ADS may result in a higher number of inconsistent behaviours. The number of datasets may affect the performance of the *DeepGuard*. However, this issue can be resolved by considering a large amount of training data during the training of *DeepGuard*.

Another core limitation of *DeepGuard* is that we tested it only on three ADS models. We also considered only three tracks (*i.e., lake, jungle, and mountain)* provided by the Udacity simulators. However, the number of inconsistent behaviours exhibited by ADSs in the urban driving environment may be higher than those considered in testing. We tried to minimize this limitation by considering only real-world best-performing ADS models (i.e., *EPOCH, CHAUFFEUR, and DAVE-2*).

## V. Conclusion

We studied autoencoders and time series analysis–based inconsistent behaviour prevention systems of DNN-based ADSs in response to the unexpected driving contexts. The prediction system of our *DeepGuard* approach outperforms the *SELFORACLE* and *DeepRoad* in terms of predictability with a high true positive rate and a very low false-positive rate in many inconsistent behavioural scenarios such as collisions and the vehicle no longer following the center of the lane. In contrast to both the *DeepRoad* and *SELFORACLE* approaches, our healing approach activates a predefined set of safety guards in advance to prevent collisions of vehicles with the roadside. In our approach, safety guards are activated before inconsistent behaviour has occurred. Safety guards act as barriers to potential collisions.

Our future work will focus on generalizing the *DeepGuard* approach to predict potential inconsistent behaviour in the urban driving scenario, as well as designing more generalized safety guards. We are able to predict inconsistent behaviour of ADS on which the *DeepGuard* is attached. However, we are also interested in predicting the inconsistent behaviour yield by front cars on highways. More specifically, in platooning driving systems, the inconsistent behaviour of vehicles may lead to the collision of the entire platooning system. Therefore, our future work will also focus on predicting the inconsistent behaviour of the front car in the platooning system and activating the safety guards to prevent collisions.

## Acknowledgement

This work is supported by the Korea National Research Foundation of the Ministry of ICT and Science (NRF2020R1A2C1007571).